# Intent-aware Multi-agent Reinforcement Learning

Siyuan Qi[1] and Song-Chun Zhu[1]

*Abstract*— This paper proposes an intent-aware multi-agent planning framework as well as a learning algorithm. Under this framework, an agent plans in the goal space to maximize the expected utility. The planning process takes the belief of other agents' intents into consideration. Instead of formulating the learning problem as a partially observable Markov decision process (POMDP), we propose a simple but effective linear function approximation of the utility function. It is based on the observation that for humans, other people's intents will pose an influence on our utility for a goal. The proposed framework has several major advantages: i) it is computationally feasible and guaranteed to converge. ii) It can easily integrate existing intent prediction and low-level planning algorithms. iii) It does not suffer from sparse feedbacks in the action space. We experiment our algorithm in a real-world problem that is non-episodic, and the number of agents and goals can vary over time. Our algorithm is trained in a scene in which aerial robots and humans interact, and tested in a novel scene with a different environment. Experimental results show that our algorithm achieves the best performance and human-like behaviors emerge during the dynamic process.

## I. INTRODUCTION

The success of human species could be attributed to our remarkable adaptability to both the physical world and the social environment. Human social intelligence endows us the ability to reason about the state of mind of other agents, and this mental state reasoning widely influences decisions made in our daily lives. For example, driving safely requires us to reason about the intents of other drivers and make decisions accordingly. This kind of subtle intent-aware decision-making (theory of mind) behavior is ubiquitous in human activities, but virtually absent from even the most advanced artificial intelligence and robotic systems.

Intent-aware decision making is particularly useful in multi-agent systems, which can find applications in a wide variety of domains including robotic teams, distributed control, collaborative decision support systems, etc. Ideally, rather than being pre-programmed with intent-aware planning behaviors, artificial intelligent agents should be able to learn the behavior in a way similar to human social adaptation. This is because designing a good behavior is difficult or even impossible, and multi-agent environments are highly non-stationary.

Fortunately, advances in machine learning and robotics algorithms provide powerful tools for solving this learning and planning problem. To design an appropriate framework for learning agents to plan upon beliefs about of agents' mental states, the following aspects need to be considered:

[1] The authors are affiliated with University of California, Los Angeles.
*This research was supported by grants DARPA XAI project N66001-17-2-4029, ONR MURI project N00014- 16-1-2007, and NSF IIS-1423305.

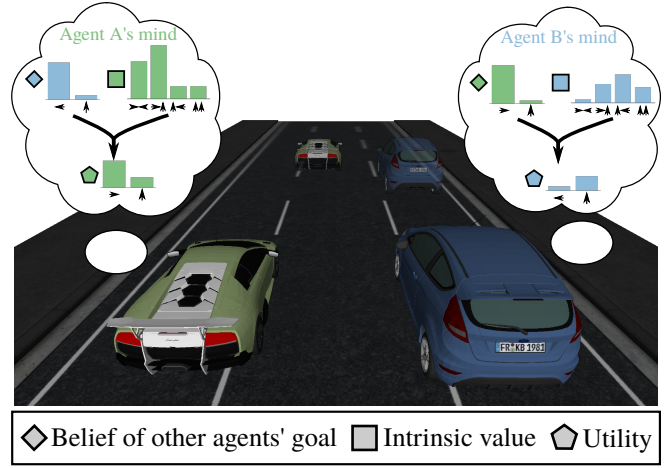

Fig. 1: Theory of mind reasoning. Safe driving requires prediction of other drivers' intent. The utility of goals are inferred based on the belief and intrinsic values, and decisions are made to maximize utility. In the figure, both drivers drove slightly across the line hence they both believe the other driver is planning to change lane. The left driver is more aggressive, having a different intrinsic value than the right one. Finally, the left driver chose to change lane since the utility is higher while the other one chose to go straight.

- Among the desired properties of a learning algorithm (e.g., guaranteed convergence, explainability, computational feasibility, fast training, ease of implementation, the degree of awareness of other agents, adaptability to non-stationary environments), which ones are more important for intent-aware planning?
- How should the framework unify the learning and planning process so that existing and future intent prediction algorithms and low-level planning methods can be fully exploited?
- The space of combined strategies of all agents can be huge. What are the important factors in the decision making process? Particularly, how important are each goal itself? How would an intent-aware agent's strategy be influenced by the goals of other agents?
- How reasonable are the learning outcomes? Will human-like social behavior (e.g., cooperations and competitions) emerge during this dynamic process?

In this paper, we propose an intent-aware planning framework as well as a learning algorithm. We propose to plan in the goal space based on the belief of other agents' intents. The intuitive idea of intent-aware planning is illustrated in

Fig. 1, in which an aggressive driver (left) and a mild driver (right) are driving on a three-lane road, and they are intended to change lane at the same time. First, both drivers infer each other's intent and based on their intrinsic values for different possibilities of strategy combinations; an expected utility is computed for their own goal. A goal is then chosen to maximize their own utility, and low-level planners are utilized to find actions to achieve the goal.

The proposed framework and algorithm provides a temporal abstraction that decouples the low-level planning, intent prediction, and high-level reasoning. The framework brings the following advantages: i) from a planning perspective, different intent prediction algorithms and low-level planners can be easily integrated. ii) By decoupling the belief update process and the learning process as opposed to POMDP, learning becomes computational feasible. iii) The temporal abstraction of planning in a goal space avoids the problem of sparse feedbacks in the action space. iv) Since any intent prediction algorithm can be adopted in this framework, no assumption is made about other agents' behaviors. The belief can be updated by various computational methods such as Bayesian methods and maximum likelihood estimation or even by communication.

The rest of the paper is organized as follows. In sec. II we review some related literature. We formally introduce the proposed planning framework in sec. III and the proposed learning algorithm in sec. IV. A real-world problem and our solution under the proposed framework is described in sec. V. We then describe the designed comparative experiment and analyze the results. Finally, sec. VI concludes the paper.

## II. RELATED WORK

*a) Intent prediction:* autonomous systems in multi-agent environments could benefit from understandings of other agents' behavior. There is a growing interest in the robotics and computer vision community to predict future activities of humans. [1], [2], [3], [4], [5], [6], [7] predict human trajectories/activities in various settings including complex indoor/outdoor scenes and crowded spaces. These research advances are potentially applicable in many domains such as assistive robotics, robot coordination, and social robotics. However, it remains unclear how these prediction algorithms can be effectively utilized for robot planning in general.

*b) Predictive multi-agent systems:* Increasing efforts have been made to design systems that are capable of predicting other agents' intents/actions to some level. Prediction algorithms have been explicitly or implicitly applied to problems such as navigation in crowds and traffic [8], [9], [10], motion planning [11], [12], and human-robot collaborative planning [13], [14]. Despite the promising results on specific problems, there lacks a framework to unify learning, prediction, and planning in general multi-agent systems.

*c) Multi-agent Reinforcement learning (MARL):* A variety of MARL algorithms have been proposed to accomplish tasks without pre-programmed agent behaviors. [15] classified MARL algorithms by their field of origin: temporal-difference RL [16], [17], [18], [19], game theory [20], [21], and direct policy search techniques [22], [23]. The degree of awareness of other learning agents exhibited by MARL algorithms depends on the learning goal. Some algorithms are independent of other learning agents and focus on convergence, while some consider adaptation to the other agents. Many deep multi-agent reinforcement learning algorithms [24], [25], for example, are agnostic of the intention of other agents. In this paper, we explicitly model other agents' intentions during the decision making process.

To model other agents' policies, POMDP is usually adopted [26], [27]. However, POMDP requires the model of a multi-agent world, and it is computationally infeasible as further discussed in Sec. III-A. We believe that by decoupling the intent inference and the learning process, we can keep the power of prediction algorithms while making the learning algorithm computational feasible.

## III. INTENT-AWARE HIERARCHICAL PLANNING

We propose an intent-aware hierarchical planning framework in multi-agent environments that plans in the goal space. The planning process is based on the belief/prediction of other agents' intents/goals inferred from the observation history. Existing low-level planners (*e.g.*, trajectory planners) are then utilized for action planning to achieve the chosen goal. In some literature, the goals are called "macro-actions".

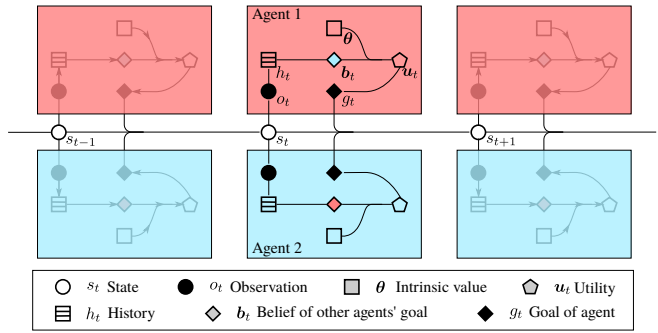

Fig. 2: The computational framework. At each time step $t$, an observation $o_t$ is made by an agent and added to history $h_t$. The intents/goals of other agents are inferred from history as belief $b_t$. Based on $b_t$ and the intrinsic value $\theta$, the utility vector $u_t$ for all goals is computed. Finally, a goal $g_t$ is choosen to maximize the utility.

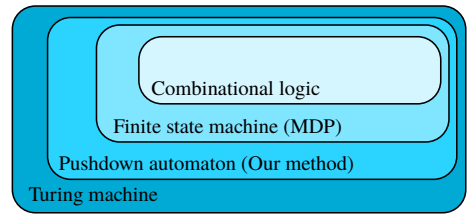

Fig. 3: Chomsky hierarchy in formal language theory.

Specifically, at each time step $t$, an agent in an environment makes an observation $o_t$ from the environment state $s_t$. The past observations are summarized into a history $h_t$. According to $h_t$, the agent infers the intent of other agents as probability distributions $\boldsymbol{b}_t$, which we call belief. A utility vector $\boldsymbol{u}_t$ is then computed for all possible goals $G$ based on $\boldsymbol{b}_t$ and the agent's intrinsic value $\boldsymbol{\theta}$. $\boldsymbol{\theta}$ is a time-invariant parameter that encodes the agent's values for different goals, and how the values would change conditioned on other agents' intent. Finally, a goal $g_t \in G$ is choosen to maximize the utility. The computational framework is illustrated in Figure 2.

This framework is motivated and inspired mainly by two schools of thoughts. Motivated by automata theory, we store the observation history $h$; inspired by theory of mind, we infer other agents' mental state and reason on the goal level.

*a) Automata theory:* the study of abstract machines and automata. It is a theory in theoretical computer science and closely related to formal language theory. An automaton is a finite representation of a formal language that may be an infinite set. Automata are often classified by the class of formal languages they can recognize, typically illustrated by the *Chomsky hierarchy* which describes the relations between various languages as shown in Figure 3. According to this classification, traditional MDPs provide a finite state machine solution to planning problems. Keeping a history $h$ of past observations makes the agent a pushdown automaton that goes beyond the finite state machine level. This frees the agent from the Markovian assumption that limits the problems in a constrained space.

*b) Theory of mind (ToM):* the intuitive grasp that humans have of our own and other people's mental state - beliefs, intents, desires, knowledge, etc. Although no one has direct access to the mind of others, it is typically assumed by humans that these mental states guide their actions in predictable ways to achieve their desires according to their beliefs. A growing body of evidence has shown that since the young-infant phase, the ability to perform mental simulations of others increases rapidly [28], [29], [30], [31]. This allows us to reason backward what others are thinking/intended for given their behavior. This reasoning process is vital in a multi-agent environment since each agent's choice affects the payoff of other agents [32], [33].

## A. A POMDP formulation and its drawbacks

For intelligent agents to adopt the proposed framework, a key facet is the learning process in which agents learn how to achieve their goals or maximize their utilities through interactions. As a result of a learning process within a population, conventions including cooperations and competitions can emerge dynamically. It is possible to formulate the learning problem as a POMDP in the belief space:

$$Q(a, \boldsymbol{b}) = \sum_s \boldsymbol{b}(s) R(s, a) + \gamma \sum_o p(o|\boldsymbol{b}, a) V(\boldsymbol{b}_o^a(s')) \quad (1)$$

$$v(\boldsymbol{b}) = \max_a Q(a, \boldsymbol{b}) \quad (2)$$

where $Q(a, b)$ denotes the long term return of taking an action $a \in A$ given the belief $b$ of other agents' intents. $R(s, a)$ denotes the immediate reward function for taking action $a$ in state $s$. A goal $g$ can be then chosen based on the return:

$$Q(g, \boldsymbol{b}) = \sum_a p(a|g) Q(a, \boldsymbol{b}) \quad (3)$$

However, this POMDP approach has two major limitations: (i) a model of the world (i.e., the underlying transition probability $p(s'|s, a)$) is required for most existing algorithms to update the belief and find the solution of the Bellman equation 1 and 2. In most cases, the agent cannot acquire the model of the world, especially in a multi-agent environment. (ii) Even for model-based methods, solving a POMDP (which can be converted to a continuous MDP) is generally computationally infeasible.

To overcome these difficulties, we propose to use an intrinsic value $\boldsymbol{\theta}$ to parametrize the utility function $u(g|\boldsymbol{b}_{-i}; h, \boldsymbol{\theta})$ as a function approximation of $Q(g, \boldsymbol{b})$ in the reinforcement learning context. This function is designed to encode the value for different combinations of agents' intents and to be computational feasible. The learning method is discussed in detail in the next section.

## IV. INTENT-AWARE MULTI-AGENT REINFORCEMENT LEARNING

Under the intent-aware hierarchical planning framework, a rational agent models uncertainty of the world and other agents via expected values of variables, and always chooses to pursue the goal $g$ with the optimal expected utility $u$ among all feasible goals:

$$g_i = \underset{g}{\operatorname{argmax}}\, u(g|\boldsymbol{b}_{-i}; h, \boldsymbol{\theta}) \quad (4)$$

where $g_i$ denotes the goal chosen by agent $i$, $\boldsymbol{b}_{-i}$ denotes the belief of intents of all the other agents, which is denoted by $-i$.

The utility function $u(g|\boldsymbol{b}_{-i}; h, \boldsymbol{\theta})$ is parameterized by the intrinsic value $\boldsymbol{\theta}$ to encode the utility for agent $i$ under different combinations of intents of all the agents. In a reinforcement learning context, the utility function $u(g|\boldsymbol{b}_{-i}; h, \boldsymbol{\theta})$ can be interpreted as a function approximation of the q-value function $Q(g, \boldsymbol{b})$.

## A. Utility function

Intuitively, the utility function should be designed in a way that the utility of all possible combinations of intents of all agents can be evaluated. A possible way to parameterize $u(g|\boldsymbol{b}_{-i}; h, \boldsymbol{\theta})$ is using a matrix $\boldsymbol{\theta}$ to represent the long-term value for all intent combinations. However, this approach is computationally inefficient to find the expected utility given the belief of other agents' intent: i) a joint probability needs to be computed for all intent combinations; ii) marginal probabilities need to be computed by summing up the joint probabilities if some agents are absent or unobservable.

Instead of directly encoding a value matrix, we use a matrix $\boldsymbol{\theta}$ to represent the influence on the value of pursuing a goal given the belief of another agent's intent. Specifically, the intrinsic value of an agent $i$ is represented by a matrix $\boldsymbol{\theta} = (\theta_{ik,jl}) \in \mathbb{R}^{(m \times n \times m)}$, where $m$ is the number of goals and $n$ is the number of agents. $\theta_{ik,jl}$ is the utility for agent $i$ to pursue goal $g_k$ if agent $j$ is intended for goal $g_l$. Then the agent choose a goal to maximize the expected utility:

$$\begin{aligned} u(g_{ik}|\boldsymbol{b}_{-i}; h, \boldsymbol{\theta}) &= \sum_j \sum_l \theta_{ik,jl} p(g_{jl}|h) \\ &= \underbrace{\theta_{ik,ik}}_{\text{agent's value}} + \underbrace{\sum_{j \neq i} \sum_l \theta_{ik,jl} p(g_{jl}|h)}_{\text{influence by other agents}} \end{aligned} \quad (5)$$

where $g_{jl}$ denotes the event of agent $j$ pursuing goal $g_l$, and $p(g_{jl}|h)$ is the corresponding probability given history. When computing the utility for $g_{ik}$, we have $p(g_{ik}|h) = 1$ and $p(g_{iq}|h) = 0$ for $q \neq k$. Hence the utility can be decomposed into two parts: $\theta_{ik,ik}$ can be interpreted as agent $i$'s value for goal $g_k$ itself, and $\sum_{j \neq i} \sum_l \theta_{ik,jl} p(g_{jl}|h)$ is the total value influence of all the other (observable) agents.

This utility function provides a linear approximation to the q-value function in a reinforcement learning setting:

$$Q(h, g_{ik}) \approx u(g_{ik}|\boldsymbol{b}_{-i}; h, \boldsymbol{\theta}) = <\boldsymbol{\theta}, \boldsymbol{\phi}_{ik}(h)> \quad (6)$$

where $\boldsymbol{\phi}_{ik}(h)$ is a feature vector that each element is given by $\phi_{j,l}(h) = p(g_{jl}|h)$, with $p(g_{ik}|h) = 1$ and $p(g_{iq}|h) = 0$ for $q \neq k$.

### B. Explainability and Theoretical Intuition

The proposed utility function provides a highly explainable model for learning in multi-agent systems. Highly complex relationships among multiple agents can arise by capturing 5 basic types of pair-wise relationships between agents:

1) Cooperation by achieving the same goal can be characterized by a positive value of $\theta_{ik,jk}$.
2) Cooperation by achieving different goals, which possibly includes temporal information, can be characterized by a positive value of $\theta_{ik,jl}$ where $k \neq l$.
3) Competition on the same goal can be characterized by a negative value of $\theta_{ik,jk}$.
4) Competition by pursuing different goals can be characterized by a negative value of $\theta_{ik,jl}$ where $k \neq l$.
5) A negligible value $|\theta_{ik,jl}| < \epsilon$ indicates $g_{jl}$ has no effect on $g_{ik}$.

From a theotical point of view, the optimal t-step POMDP value function is proven to be piecewise linear and convex. In our approximation, the value function $v(\boldsymbol{b}) = \max_g Q(h, g) \approx \max_{g_{ik}} <\boldsymbol{\theta}, \boldsymbol{\phi}_{ik}(h)>$ is also piecewise linear and convex. Hence this utility function provides a reasonable approximation of the POMDP solution given by Eq. 2.

Another reason that the proposed utility function is well suited for reinforcement learning is that it provides desired convergence properties for learning. Linear function approximation is known to be able to learn efficiently from incrementally acquired data and guaranteed to converge in different problem settings, including both episodic and continuing cases [34]. Non-stationary target functions (target functions that change over time) can also be handled for generalized policy iteration. Hence different learning algorithms can be applied, including bootstrapping/non-bootstrapping and on-policy/off-policy algorithms.

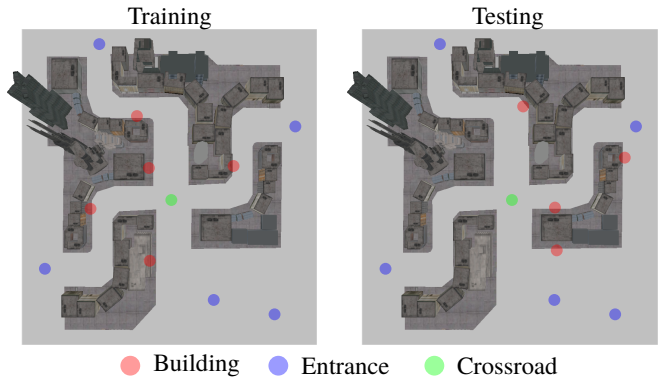

Fig. 4: Experiment settings of multi aerial robot surveillance. Left: the training scene. Right: the testing scene. At any time frame, each of the aerial robots can choose to monitor one of the targets in the goal space: buildings, entrance points to the town, and crossroads. Humans enter the town from entrance points randomly and either enter a building or exit from an entrance point. The goal of the aerial robots is to capture as many as possible people that enter the buildings of interest.

### C. Reachable Equilibrium

In a multi-agent environment, one important aspect to consider for learning agents is whether the learning algorithm will reach a meaningful *solution concept*, which identifies a certain subset of outcomes in a game. Among different solution concepts, *Nash equilibrium* is the most influential one describing the situation that no agent would want to change his/her strategy if he/she knew what strategies the other agents were following. In the proposed framework, the prediction of other agents' intent provides a one-step look ahead of their strategies. At each time step, the agent makes a game theoretic *best response* based on his/her belief of other agents' strategy. Hence a Nash equilibrium will be reached by definition when the agents' decisions stabilize.

However, whether the agents' decisions will stabilize is a subtle issue. Consider the case when two agents on opposite sides of a bridge and both agents want to pass through. The interaction process can easily go into a deadlock that an equilibrium cannot be reached: at time $t_0$ both agents choose to go, and at time $t_1$ both agents choose to yield, etc. Computationally, having asynchronous decision-making agents can solve this problem. Instead of making a new decision at each time step, an agent chooses to keep the previous decision according to a certain probability. At some point, the asynchronous agents will break the deadlock and reach an equilibrium.

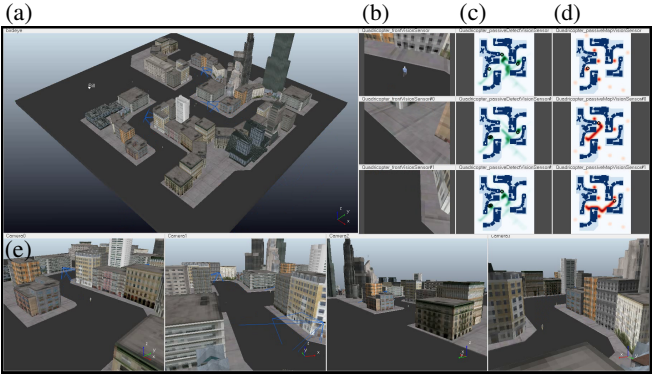

Fig. 5: The simulation environment. The viewing frustums of aerial robots are rendered in blue lines. (a) Birdview of the city scene. Each row of (b)(c)(d) shows the vision and mind of an aerial robot agent: (b) vision sensor input of an aerial robot, (c) intent and path prediction of other agents, and (d) current goal of the aerial robot. (e) Other views from street cameras.

## V. Multi Aerial Robot Surveillance

One of many motivations of our research effort on intent-aware multi-agent reinforcement learning is the application of multi-aerial robot surveillance, in which each robot needs to infer both the humans' intents and teammate robots' intents to accomplish their tasks. Humans' intents need to be inferred to capture suspicious activities, while other robots' intents need to be inferred to maximize the payoff of the team (e.g., by avoiding repeating other robots' actions). This multi-agent environment is neither purely cooperative nor competitive, and the goal is to learn a strategy for aerial robots to perform the task distributively.

Specifically, we experiment our algorithm in the following setting as shown in Fig. 4. In a city area, humans can enter the scene randomly from entrance points of the area, and either exit the scene from other entrance points or enter a building of interest. In other words, the goals for humans are entrance points or buildings. We are interested in finding out the people who enter the buildings of interest using a team of aerial robots, each of which is capable of monitoring a static target or tracking a moving target. Each time an aerial robot's camera sees a human entering a building, we consider it to have captured one suspicious human.

For a city scene, we define three types of static targets/goal: entrance points, buildings, and crossroads. At each time step, a robot can choose to monitor a static target or track an observed human. A given path planner will drive the robot to the target after the decision.

### A. Method

In this subsection, we describe our method for the task above under the intent-aware reinforcement learning framework. Specifically, we describe i) the learning algorithm adopted for estimating the parameter $\boldsymbol{\theta}$, ii) how to deal with a varying number of agents and goals, and iii) how to predict the intents of other agents in this setting.

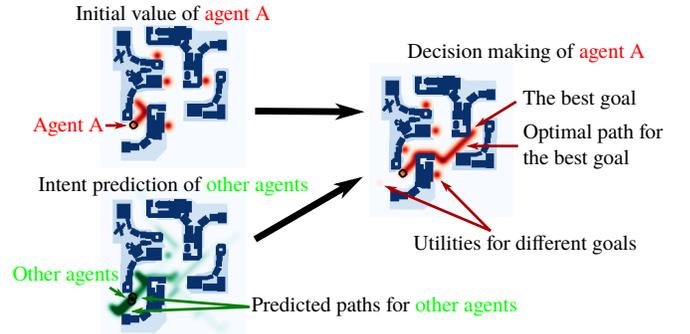

Fig. 6: Decision making process of an intent-aware agent in the simulation environment. The agents' positions are represented by black circles.

*a) Average award learning:* the problem described above is a continuing problem (non-episodic) that the discounted reward setting does not apply. To learn a strategy for each aerial robot, we adopt the differential semi-gradient Sarsa algorithm to learn the parameter $\boldsymbol{\theta}$, where the temporal-difference (TD) error to update $\boldsymbol{\theta}$ is given by the estimation error of the average reward [35]

$$\delta = r_{t+1} - \bar{r}_t + Q(h_{t+1}, g_{t+1}; \boldsymbol{\theta}) - Q(h_t, g_t; \boldsymbol{\theta}) \quad (7)$$

where $\bar{r}$ is an estimate of the average reward. At time $t+1$, the parameter $\boldsymbol{\theta}$ and the average reward estimation $\bar{r}$ is updated by:

$$\boldsymbol{\theta}_{t+1} = \boldsymbol{\theta}_{t+1} + \alpha \delta \nabla Q(h_t, g_t; \boldsymbol{\theta}_t) \quad (8)$$

$$\bar{r}_{t+1} = \bar{r}_t + \beta \delta \quad (9)$$

where $\alpha$ and $\beta$ are learning rates. Algorithm 1 shows the 1-step differential semi-gradient Sarsa for our problem, which can be generalized to n-step bootstrapping.

*b) Typed agents and goals:* another difficulty for solving this problem is that at different time steps, an agent can have a different number of agents to interact with and various number of goals to choose from, e.g., the robot observes more people at some point. This sets an indefinite size for $\boldsymbol{\theta}$. To handle this, we define $\boldsymbol{\theta}$ in a compact way. Each element $\theta_{t(i,k,j,l)}$ represents the value influence of an agent of type $t(j)$ choosing a type of goal $g_l$ on the event that agent of type $t(i)$ choosing a type of goal $g_k$. Here $t$ is a mapping function from agents and goals to the corresponding types. Now the q-function becomes:

$$Q(h, g_{ik}) = \sum_j \sum_l \theta_{t(i,k,j,l)} p(g_{jl}|h) \quad (10)$$

*c) Intent prediction:* during the interaction process, we use a Bayesian approach to predict other agents' intents $g$ based on observation history $h$:

$$p(g|h) = \frac{p(g)p(h|g)}{p(h)} \quad (11)$$
$$\propto p(g)p(h|g)$$

where $p(g)$ is a prior probability given by the agent's own q-value estimation of the goal or a uniform distribution. $p(h|g)$

**Algorithm 1:** Intent-aware learning in continuing tasks

**Input**: Linear approximation function $Q(h, g; \boldsymbol{\theta}) = <\boldsymbol{\theta}, \phi(h)>$
**Parameters:** Learning rate $\alpha, \beta > 0$
Goal update frequency $f$

1. Initialize $\boldsymbol{\theta}, \bar{r}$ arbitrarily (e.g., $\boldsymbol{\theta} = \mathbf{0}, \bar{r} = 0$)
2. Initialize obervation $o$ and goal $g$
3. Initialize history $h$ as an empty stack
4. **while** *True* **do**
5.    Plan for $g$, observe $r$, $o$
6.    Push $o$ into $h$: $h' \leftarrow [h, o]$
7.    **if** $r == 0$ *and* $random() < \frac{1}{f}$ **then**
8.      // Keep goal with randomness
     **continue**
9.    **else**
10.      Compute $\phi(h, o)$ by inferring intent of other agents
11.      Choose goal $g'$ as a function of $Q(h, \cdot; \boldsymbol{\theta})$ (e.g., $\epsilon$-greedy)
12.      **if** $r! = 0$ **then**
13.        $\delta \leftarrow r - \bar{r} + Q(h', g'; \boldsymbol{\theta}) - Q(h, g; \boldsymbol{\theta})$
14.        $\bar{r} \leftarrow \bar{r} + \beta\delta$
15.        $\boldsymbol{\theta} \leftarrow \boldsymbol{\theta} + \alpha\delta\nabla Q(h, g; \boldsymbol{\theta})$
16.      **end**
17.      $o \leftarrow o', g \leftarrow g'$
18.    **end**
19. **end**

TABLE I: Surveillance capture rate

| | Testing Accuracy(%) | | | | | |
|---|---|---|---|---|---|---|
| | Training Scene | | | | | Testing Scene |
| Training Iterations | 100 | 200 | 300 | 500 | 1000 | — |
| random | 29.7 | 29.7 | 29.7 | 29.7 | 29.7 | 34.4 |
| greedy | **31.4** | **31.4** | 31.4 | 31.4 | 31.4 | 35.2 |
| RNN-POMDP | 17.3 | 22.9 | 20.8 | 21.0 | 23.6 | 14.3 |
| ours | 13.8 | 19.1 | **60.3** | **62.4** | **59.3** | **77.3** |

Models trained for 1000 iterations in the training scene are used for testing in the testing scene.

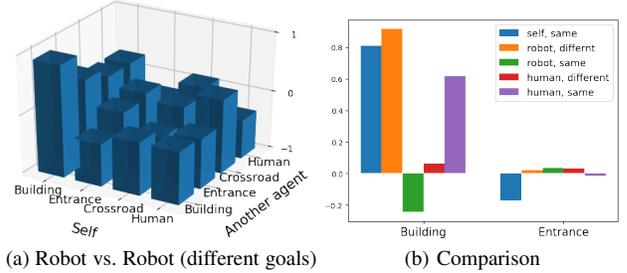

(a) Robot vs. Robot (different goals)     (b) Comparison

Fig. 7: Visualization of the learned value landscape. (a) The value influence by other robots if they are intended for a different goal (the goal types can be the same) than the agent robot. Regardless of what the other robots are intended to, the value for monitoring buildings is the highest. (b) Comparison between the value influences from different types of agents. From this comparison we see that: i) a robot agent has a positive value for monitoring a building while having a negative value for monitoring an entrance point. ii) The value for two robots to monitor different buildings is high while there is a penalty for monitoring the same buildings. iii) There is a value for a robot to monitor the same building a human is intended to. iv) Intents of other agents have almost no influence on the value of monitoring an entrance point.

is a likelihood term that measures how likely the observation history is if the agent is pursuing goal $g$. It is given by the following Gibbs distribution:

$$p(h|g) = \frac{1}{z} \exp\{-\beta E(h|g)\}$$
$$= \frac{1}{z} \exp\{-\beta d(\Gamma_o, \Gamma_p)\} \quad (12)$$

where the engergy function $E(h|g)$ is defined by a distance $d(\Gamma_o, \Gamma_p)$. $\Gamma_o$ is the observed trajectory from history $h$, and $\Gamma_p$ is a predicted trajectory assuming the agent is heading for goal $g$. We compute the distance $d(\Gamma_o, \Gamma_p)$ between $\Gamma_o$ and $\Gamma_p$ by the dynamic time warping (DTW) algorithm [36]. The intuition is that the closer the observed trajectory is to the predicted one, the higher the probability is for goal $g$. $\beta$ is the inverse temperature that controls the distribution landscape of the distribution. $z$ is the partition function of the Gibbs distribution to normalize the probability to 1.

*B. Simulation Environment*

For the robot agents, the goal space is the set of the previously defined targets (humans, buildings, and entrances); the action space includes: forward, backward, left, and right. The observations of other agents are their positions at each time step. The state is composed of the goal and position of the agent itself and the observation of other agents.

We simulate the environment in V-REP [37] as shown in Fig. 5. The vision sensor inputs of aerial robots are published to ROS topics [38], and humans are detected using the YOLO [39] object detector. Once a target is given to an aerial robot, a path is planned to monitor/track the target using [40]. In the experiments, we assume that the aerial robots know each other's position but not their goals. They can compute an observed human's position and share to other robots. The targeted buildings are known to the robot agents prior to the start of the simulation. At each simulation time step, a state including all observable agents' positions will be sent to robot agents. A reward will be sent to robot agents if a human agent enters a building. A reward of $\frac{1}{m}$ will be assigned to the $m$ robots who observe the entering. A reward of $-\frac{1}{n}$ will be assigned to all $n$ robot agents if a human enters a building unobserved. In our experiments, three aerial robot agents are simulated in the experiments, and human agents enter the scene randomly. The training scene has 5 buildings of interest while the testing scene has 4, and the locations are different.

*C. Comparative Methods*

We compare our method against the following baselines:
1) Random. The robot agent chooses a random goal to monitor/track after a certain number of simulation iterations.

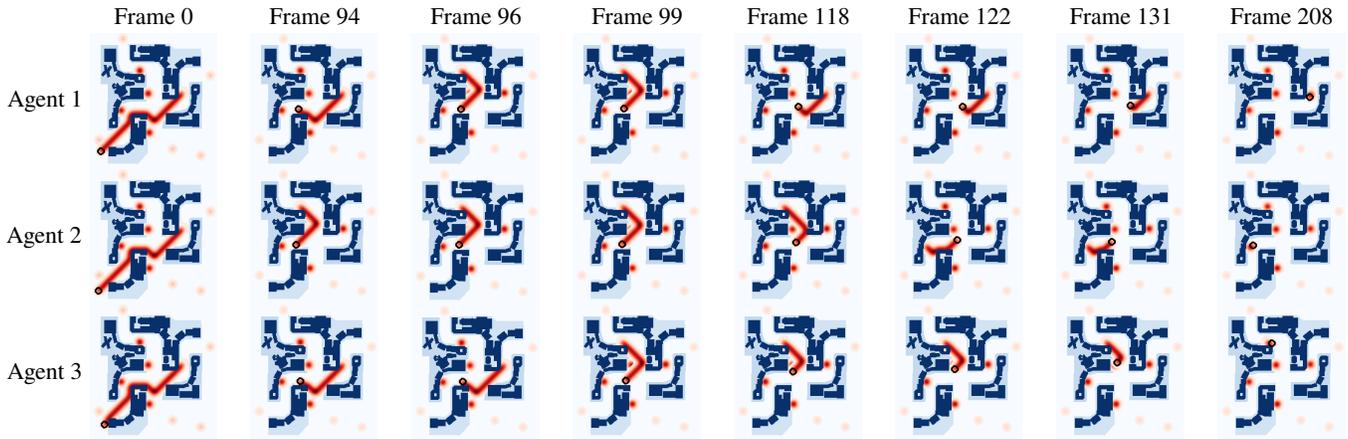

Fig. 8: Cooperative behavior of the learned model. Each row demonstrates the values of different goals (the higher the value the higher intensity of red), the chosen goal, and the planned path for a robot at different time frames. The values of monitoring buildings are higher than monitoring entrances or crossroad. At the beginning, all robot agents intended for the farthest building since they thought it is less likely for the other robots to monitor. During the process, the agents realized that the other agents might have chosen the same goal as themselves and chose another goal accordingly (e.g., agent 2 at frame 94). The agents chose different goals iteratively and finally reached an equilibrium that three robots were monitoring three different buildings. This dynamic process reflects how humans naturally interact with each other in a complex environment.

2) *Greedy.* The robot tracks humans whenever a human is observed. Otherwise, a goal will be chosen at random.

3) *Recurrent neural network (RNN)-POMDP [41].* Since the agents have no access to the underlying model of the world, a model-free method is needed to compare our method against POMDP algorithms. A RNN-POMDP agent directly approximates the Q function by

$$Q(g, o) = f(o, g; W) \quad (13)$$

where $W$ is the weight in the RNN. In our experiments, we use the LSTM(512)-FC(128)-FC(32) architecture, where FC($n$) represents a fully connected layer with $n$ hidden nodes. In order to prevent the network from diverging (numeric explosion of weights), we treat "15 people have entered buildings" as an episode, and use Sarsa to train the network with an $\epsilon$-greedy policy.

For both RNN-POMDP and our method, the robot agents share the learned policy during the training stage. Tan [42] has shown that reinforcement learning agents can learn faster and converge sooner in this way.

### D. Experiment Results & Discussions

Fig. 6 demonstrates the decision-making process of our agents, and Table I shows the quantitative results of our experiments. The evaluation metric is defined by $\frac{n_o}{n_e}$ where $n_e$ is the total number of people that entered buildings and $n_o$ is the number of observed ones. It shows the testing capture rate on the training scene of trained models after 100, 200, 300, 500 and 1000 training iterations (simulation steps that reward occurs). It also shows the testing capture rate on the testing scene using the final model.

Our method outperforms all baseline methods and achieves a significant increase in the capture rate even in a novel situation. After 300 iterations of value function update, it has converged to the global optimal. We also experimented our method with random initializations for the parameter $\boldsymbol{\theta}$ and 5-step Sarsa, and they all achieved similar performance. More importantly, our method is able to transfer the learned strategy to a novel scene and achieves the best performance.

*What is the best strategy?* The intent-aware agents eventually learned a strategy that reflects two facts: i) it is more efficient to monitor a building directly instead of tracking human agents. ii) Each robot agent needs to monitor a different building. This means our method successfully learned the underlying value as long as cooperation. A capture rate of 60% is achieved, for 3 robot agents monitoring 5 buildings at any time frame. From the baseline results, we can see that greedy algorithm performs slightly better than random, since tracking a human agent sometimes distracts the robot agent. In the experiments, we see that the RNN-POMDP agent learned to monitor a building; however, all three robot agents decided to monitor the same building according to the best approximated Q-value. Hence they achieved a capture rate of 20% (one out of five buildings).

*How meaningful is the learned $\boldsymbol{\theta}$?* Figure 7 shows a visualization of the value landscape given by $\boldsymbol{\theta}$. The intent-aware agents learned meaningful aspects including i) it is rewarding to monitor buildings, especially the ones that no other robot is monitoring. ii) Monitoring the same building with other robot agents is undesirable. iii) Monitoring a building that an observed human is targeting at is rewarding. This demonstrates that the agent learns the core values of the task, though no explicit knowledge is given. This is important for transferring the learned strategy to novel scenes.

*How does the cooperation emerge and reach equilibrium?* Figure 8 shows the interesting cooperative behavior occurred. All robot agents are well aware that it is rewarding to monitor different buildings than other agents. At the beginning, they were targeted at the farthest building since it is less likely for the other robots to monitor. After a few iterations, they realized there might be a conflict of interest and changed their mind. Finally, three robots are monitoring three different buildings. An equilibrium naturally emerged during this dynamic process, which resembles human interactions. After that, the robots sometimes switch to another un-monitored building at random for the same utility as the current one, which also matches human behaviors.

*How would the design of rewards affect the learned behaviors?* In our experiments, we found that assigning rewards to those robots who capture entering is more effective than assigning rewards to every robot in the team. Cooperation might not emerge if a reward is distributed to every robot since a robot can be credited even if it has made a wrong decision. This is surprisingly similar to human behaviors: teamwork works best when top performers are rewarded.

## VI. CONCLUSION

In this work, we consider the problem of multi-agent planning as well as learning in a goal space. The proposed planning framework infers other agents' intents and makes decisions based upon the prediction and the intrinsic value. Rather than formulating the problem as a POMDP, we decouple the intent prediction and the high-level planning process, keeping the capability of the algorithm while making the method computationally feasible. Experiments show that our algorithm achieves the best performance, and behaviors similar to humans emerges under the proposed framework. More sophisticated intent prediction and high-level planning algorithms can be developed independently in future research, and integrated into the proposed framework.